\documentclass{article}

     \usepackage[final]{neurips_2019}

\usepackage[utf8]{inputenc} 
\usepackage[T1]{fontenc}    
\usepackage{hyperref}       
\usepackage{url}            
\usepackage{booktabs}       
\usepackage{amsfonts}       
\usepackage{nicefrac}       
\usepackage{microtype}      

\usepackage{amsmath}
\usepackage{algorithm, algorithmicx}
\usepackage{algpseudocode}
\usepackage{graphicx}
\usepackage{xcolor}
\usepackage{hyperref}

\title{iDLG: Improved Deep Leakage from Gradients}

\author{%
   Bo Zhao, Konda Reddy Mopuri, Hakan Bilen 
     \\
   School of Informatics\\
   The University of Edinburgh, United Kingdom \\
   \texttt{\{bo.zhao, kmopuri, hbilen\}@ed.ac.uk} \\
}

\begin{document}

\maketitle

\begin{abstract}
It is widely believed that sharing gradients will not leak private training data in distributed learning systems such as Collaborative Learning and Federated Learning, etc. Recently, Zhu \textit{et al.}~\cite{zhu19deep} presented an approach which shows the possibility to obtain private training data from the publicly shared gradients. In their Deep Leakage from Gradient (DLG) method, they synthesize the dummy data and corresponding labels with the supervision of shared gradients. However, DLG has difficulty in convergence and discovering the ground-truth labels consistently. 
In this paper, we find that sharing gradients definitely leaks the ground-truth labels. 
We propose a simple but reliable approach to extract accurate data from the gradients. Particularly, our approach can certainly extract the ground-truth labels as opposed to DLG, hence we name it Improved DLG (iDLG). Our approach is valid for any differentiable model trained with cross-entropy loss over one-hot labels. We mathematically illustrate how our method can extract ground-truth labels from the gradients and empirically demonstrate the advantages over DLG.
\end{abstract}

\section{Introduction}
In multi-node distributed learning systems such as Collaborative Learning \cite{shokri2015privacy, song2018collaborative, melis2018inference} and Federated Learning \cite{konevcny2016federated, mcmahan2017federated, li2019federated}, it is widely believed that sharing gradients between nodes will not leak the private training data. In the popular setup, all the individual participants aim to learn a shared model in a centralized or decentralized manner. They would share the individual gradients and update the model parameters with the aggregated gradients. In these frameworks, it is a common practice to share only the gradients in order protect the proprietary data.  However, recent work by Zhu \textit{et al.}, ``Deep Leakage from Gradient'' (DLG) \cite{zhu19deep} showed the possibility to steal the private training data from the shared gradients of other participants.

The main idea of DLG is to generate dummy data and corresponding labels via matching the dummy gradients to the shared gradients. Specifically, they start with randomly initialized the dummy data and labels. From there, they compute dummy gradients over the current shared model in the distributed setup. Via minimizing the difference between dummy gradients and the shared real gradients, they iteratively update the dummy data and labels simultaneously. Although DLG works, we find that it is not able to reliably extract the ground-truth labels or generate good quality dummy data.

In this paper, we propose a simple but definitely valid approach to extract the ground-truth labels from the shared gradients. By derivation, we demonstrate that the gradient of the classification (cross-entropy) loss w.r.t. the correct label activation (in the output layer) lies in $(-1, 0)$, while those of other labels lie in $(0, 1)$. Hence, the signs of gradients w.r.t. correct and wrong labels are opposite. When the gradients w.r.t. the outputs (logits) are not accessible, we show that the gradients w.r.t. the last-layer weights (between the output layer and the layer in front of it) also follow this rule. With this rule, we can identify the ground-truth labels based on the shared gradients. In other words, the ground-truth labels are definitely leaked by sharing gradients of a Neural Network (NN) trained with cross-entropy loss. This enables us to always extract the ground-truth labels and significantly simplify the objective of DLG~\cite{zhu19deep} in order to extract good-quality data. Hence, we name our approach, Improved DLG (iDLG). The main contributions of our work includes:
\begin{itemize}
    \item By revealing the relationship between labels and signs of gradients, we present an analytical procedure to extract the ground-truth labels from the shared gradients with $100\%$ accuracy, which facilitates the data extraction with better fidelity.
    \item We empirically demonstrate the advantages of iDLG over DLG~\cite{zhu19deep} via comparing the accuracy of extracted labels and the fidelity of extracted data on three datasets.
\end{itemize}

The rest of the paper is organised as follows: Section~\ref{sec:iDLG} presents the analytical procedure to extract the ground-truth labels from the shared gradients and the proposed iDLG method. Section~\ref{sec:experiments} demonstrates the advantages of iDLG over DLG through the experimental evaluation, and Section~\ref{sec:conclusion} concludes the paper with discussion.

\section{Methodology}
\label{sec:iDLG}
Recent work by Zhu \textit{et al.}~\cite{zhu19deep} presents an approach (DLG) to steal the proprietary data protected by the participants in distributed learning from the shared gradients. In their method, they attempt to generate the dummy data and corresponding labels via a gradient matching objective. However, in practice, it is observed that their method generates wrong labels frequently. In this work, we present an analytical approach to extract the ground-truth labels from the shared gradients, then we can extract the data more effectively based on correct labels. Hence, we name our approach, improved Deep Leakage from Gradients (iDLG). In this section, we first present the procedure to extract the ground-truth labels. Then, we show the iDLG method based on the extracted labels.

\subsection{Extracting Ground-truth Labels}
Let us consider the classification scenario, where the NN model is generally trained with cross-entropy loss over one-hot labels, which is defined as
\begin{equation}
        l(\mathbf{x}, c) = -\log \frac{e^{y_c}}{\Sigma_j e^{y_j}},
\end{equation}
where $\mathbf{x}$ is the input datum, $c$ is the corresponding ground-truth label. $\mathbf{y}=[y_1, y_2, ...]$ is the outputs (logits), and $y_i$ denotes the score (confidence) predicted for the $i^{th}$ class. Then, the gradients of the loss w.r.t. each of the outputs is 
\begin{equation}
\begin{split}
{g}_i = \frac{\partial l(\mathbf{x}, c)}{\partial y_i} = & -\frac{\partial \log e^{y_c} - \partial \log \Sigma_j e^{y_j}}{\partial y_i} \\
= & \left\{ \begin{aligned}
-1 + & \frac{e^{y_i}}{\Sigma_j e^{y_j}}, \;\;\; if \;\; i = c\\
 & \frac{e^{y_i}}{\Sigma_j e^{y_j}}, \;\;\; else \\
    \end{aligned} \right.
\end{split}
\label{eqn:gi}
\end{equation}
As the probability $\frac{e^{y_i}}{\Sigma_j e^{y_j}} \in (0, 1)$, we have ${g}_i \in (-1, 0)$ when $i=c$ and ${g}_i \in (0, 1)$ when $i \neq c$. Hence, we can identify the ground-truth label as the index of the output that has the negative gradient.

However, we may not be able to access the gradients w.r.t. the outputs $\mathbf{y}$, as they are not included in the shared gradients $\nabla \mathbf{W}$ which are the derivatives w.r.t. the weights of the model $\mathbf{W}$.  
We find that the gradient vector $\nabla \mathbf{W}_L^i$ w.r.t. the weights $\mathbf{W}^i_L$ connected to the $i^{th}$ logit in the output layer can be written as
\begin{equation}
\begin{split}
\nabla \mathbf{W}_L^i = \frac{\partial l(\mathbf{x}, c)}{\partial \mathbf{W}_L^i} & =   \frac{\partial l(\mathbf{x}, c)}{\partial y_i} \cdot \frac{\partial y_i}{\partial \mathbf{W}_L^i} \\
& = {g}_i \cdot \frac{\partial ({\mathbf{W}_L^i}^T\mathbf{a}_{L-1} + b_{L}^i)}{\partial \mathbf{W}_L^i} \\
& = g_i \cdot \mathbf{a}_{L-1},
\end{split}
\end{equation}
where the network has $L$ layers, $\mathbf{y} = \mathbf{a}_L$ is the output layer activations, $b_{L}^i$ is the bias parameter, and $y_i = {\mathbf{W}_L^i}^T\mathbf{a}_{L-1} + b_{L}^i$.
As the activation vector $\mathbf{a}_{L-1}$ is independent of the class (logit) index $i$, we can easily identify the ground-truth label according to the sign of $\nabla \mathbf{W}_L^i$ which is different from others. Therefore, the ground-truth label $c$ is predicted as
\begin{equation}
c = i, \text{\:\:\: s.t. \:\:\:}  {\nabla \mathbf{W}_L^i}^T \cdot {\nabla \mathbf{W}_L^j} \leq 0, \:\:\: \forall j \neq i
\label{eqn:gt-label}
\end{equation}
When the non-negative activation function, \emph{e.g.} ReLU and Sigmoid, is used, the signs of $\nabla \mathbf{W}_L^i$ and $g_i$ are the same. Hence, we can simply identify the ground-truth label whose corresponding $\nabla \mathbf{W}_L^i$ is negative.
With this rule, it is easy to identify the ground-truth label $c$ of the private training datum $\mathbf{x}$ from shared gradients $\nabla \mathbf{W}$. Note that this rule is independent of the model architectures and parameters. In other words, this holds for any network at any training stage from any randomly initialized parameters.

\subsection{Improved DLG (iDLG)}
Based on the extracted ground-truth labels, we propose the improved DLG (iDLG) which is more stable and efficient to optimize. The algorithm is illustrated in Algorithm~\ref{alg:iDLG}. The iDLG procedure starts with the differentiable learning model $F(\mathbf{x}; \mathbf{W})$ with the model parameters $\mathbf{W}$, and the gradients $\nabla \mathbf{W}$ calculated based on private training datum $(\mathbf{x}, c)$. The first step is to extract the ground-truth label $c'$ from the shared gradients $\nabla \mathbf{W}$ as in eq~(\ref{eqn:gt-label}).

Then, we randomly initialize the dummy datum $\mathbf{x}' \xleftarrow{} \mathcal{N}(0,1)$. We calculate the dummy gradients $\nabla \mathbf{W}'$ based on the dummy datum and the extracted label $(\mathbf{x}', c')$. The training objective is to match the dummy gradients with the shared gradients, \emph{i.e.}, to minimize
 \begin{equation}
  L_G = \|\nabla \mathbf{W}' - \nabla \mathbf{W}\|^2_F      
\end{equation}

Based on this training objective, we update the dummy datum $x'$ by gradient descent
 \begin{equation}
  \mathbf{x}' \leftarrow{} \mathbf{x}' - \eta \nabla_{\mathbf{x}'} {L_G}   
\end{equation} 
for $N$ training iterations, where $\eta$ is the learning rate.

\begin{algorithm}
\small
\caption{\small{Improved Deep Leakage from Gradients  (iDLG)}\label{alg:euclid}}
\begin{algorithmic}[1]
\Require 
    \Statex $F(\mathbf{x}; \mathbf{W})$: Differentiable learning model, $\mathbf{W}$: Model parameters,  $\nabla \mathbf{W}$: Gradients produced by private training datum $(\mathbf{x}, c)$, $N$: maximum number of iterations. $\eta$: learning rate.
\Ensure 
    \Statex $(\mathbf{x}', c')$: Dummy datum and label.
\State $c' \xleftarrow{} i \text{\:\:\: s.t. \:\:\:}  {\nabla \mathbf{W}_L^i}^T \cdot {\nabla \mathbf{W}_L^j} \leq 0, \:\:\: \forall j \neq i$ \;\;\;\;\;\;\; $\triangleright$ Extract the ground-truth label.
\State $\mathbf{x}' \xleftarrow{} \mathcal{N}(0,1)$ \;\;\;\;\;\;\; $\triangleright$ Initialize the dummy datum.
\For{$i \xleftarrow{} 1$ to $N$}
    \State $\nabla \mathbf{W}' \xleftarrow{} \partial l(F(\mathbf{x}'; \mathbf{W}), c') / \partial \mathbf{W}$ \;\;\;\;\;\;\; $\triangleright$ Calculate the dummy gradients.
    \State $L_G = \|\nabla \mathbf{W}' - \nabla \mathbf{W}\|^2_F$ \;\;\;\;\;\;\; $\triangleright$ Calculate the loss (difference between gradients).
    \State $\mathbf{x}' \xleftarrow{} \mathbf{x}' - \eta \nabla_{\mathbf{x}'} {L_G}$ \;\;\;\;\;\;\; $\triangleright$ Update the dummy datum.
\EndFor
\end{algorithmic}
\label{alg:iDLG}
\end{algorithm}

\section{Experiments}
\label{sec:experiments}
In this section, we empirically demonstrate the advantages of our (iDLG) method over DLG~\cite{zhu19deep}. We perform experiments on the classification task over three datasets: MNIST~\cite{lecun1998gradient}, CIFAR-$100$~\cite{krizhevsky2009learning}, and LFW~\cite{huang2008labeled} with $10$, $100$, and $5749$ categories respectively. Following the settings in \cite{zhu19deep}, we use the randomly initialized LeNet for all experiments. L-BFGS \cite{liu1989limited} with learning rate $1$ is used as the optimizer. For fast training, we resize all images in LFW to $32\times 32$.

For DLG~\cite{zhu19deep}, as described by the authors, we start the procedure with the randomly initialized dummy data and outputs $(\mathbf{x}', \mathbf{y}')$, then iteratively update them to minimize the gradient matching objective. For both two algorithms, we perform the optimization for $300$ iterations, and evaluate the performance in terms of (i) the accuracy of the extracted labels $c'$, and (ii) the fidelity of the extracted data $\mathbf{x}'$.
We run all experiments for $1000$ times with randomly initialized networks and report the mean values. The code has been released on GitHub\footnote{\href{https://github.com/PatrickZH/Improved-Deep-Leakage-from-Gradients}{https://github.com/PatrickZH/Improved-Deep-Leakage-from-Gradients}}. 

\begin{table}[]
\centering
\begin{tabular}{lll}
\hline
\multicolumn{1}{l}{Dataset} & \multicolumn{1}{l}{DLG} & \multicolumn{1}{l}{iDLG} \\ \hline
MNIST                         & 89.9\%                      & 100.0\%                       \\
CIFAR-100                     & 83.3\%                       & 100.0\%                       \\
LFW                           & 79.1\%                     & 100.0\%   \\ \hline                   
\end{tabular}
\caption{\small{Accuracy of the extracted labels for DLG~\cite{zhu19deep} and iDLG. Note that iDLG always extracts the correct label as opposed to DLG which extracts wrong labels frequently.}}
\label{tab:label-accuracy}
\end{table}

\begin{figure}[]
\small
\centering
\noindent\begin{minipage}{\textwidth}
  \centering
  \includegraphics[width=.33\textwidth]{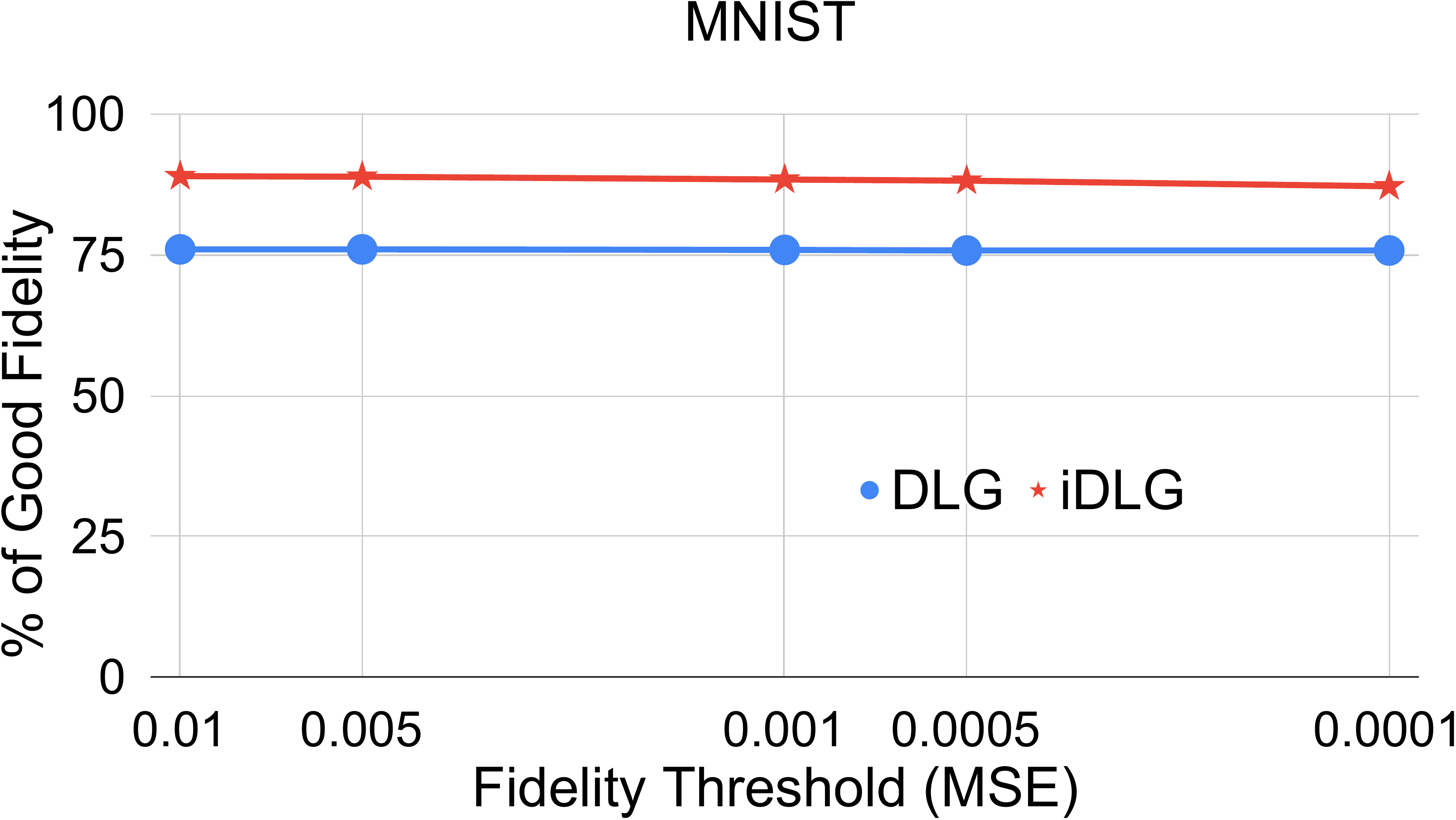}\hfill
  \includegraphics[width=.33\textwidth]{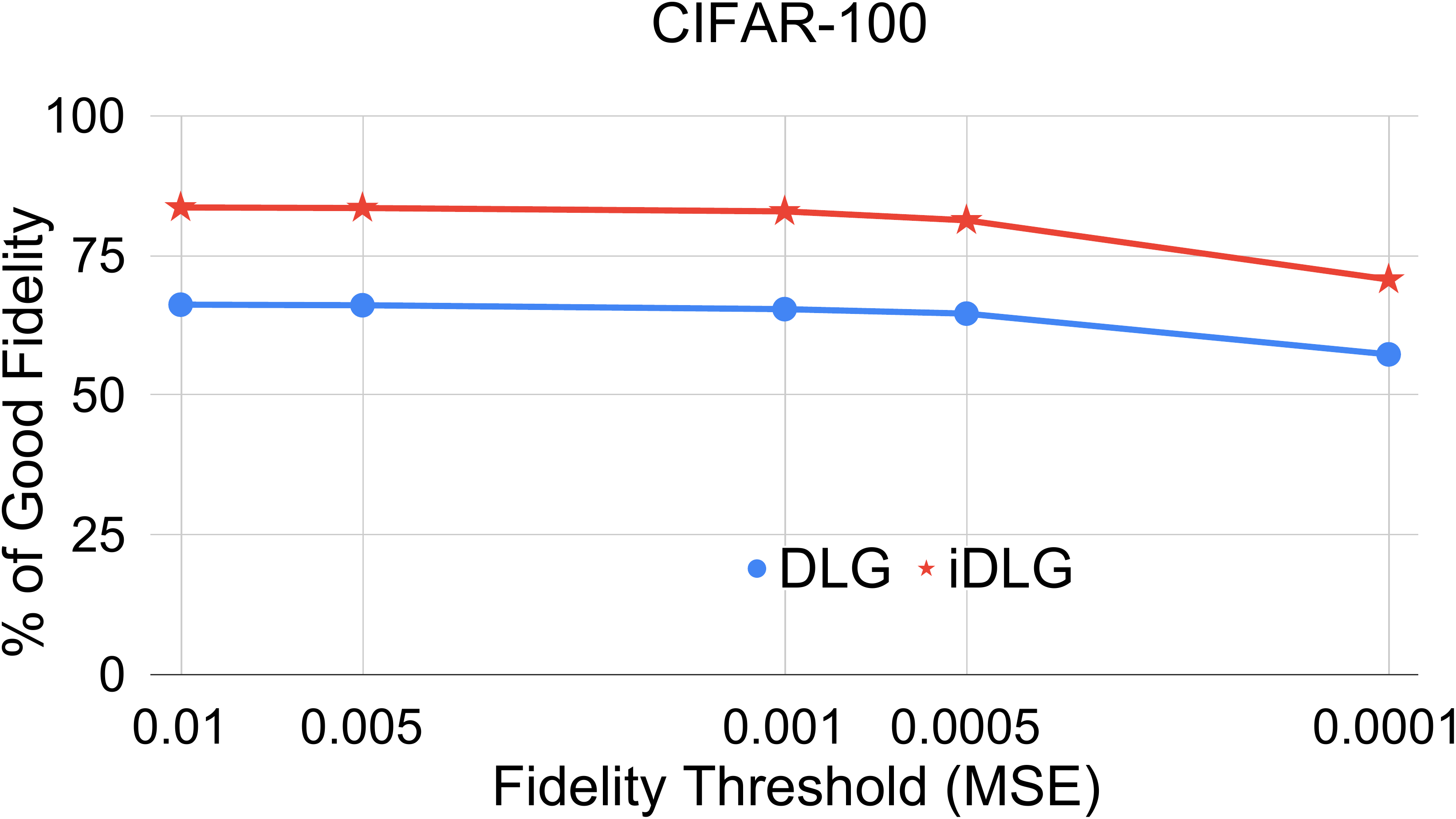}\hfill
  \includegraphics[width=.33\textwidth]{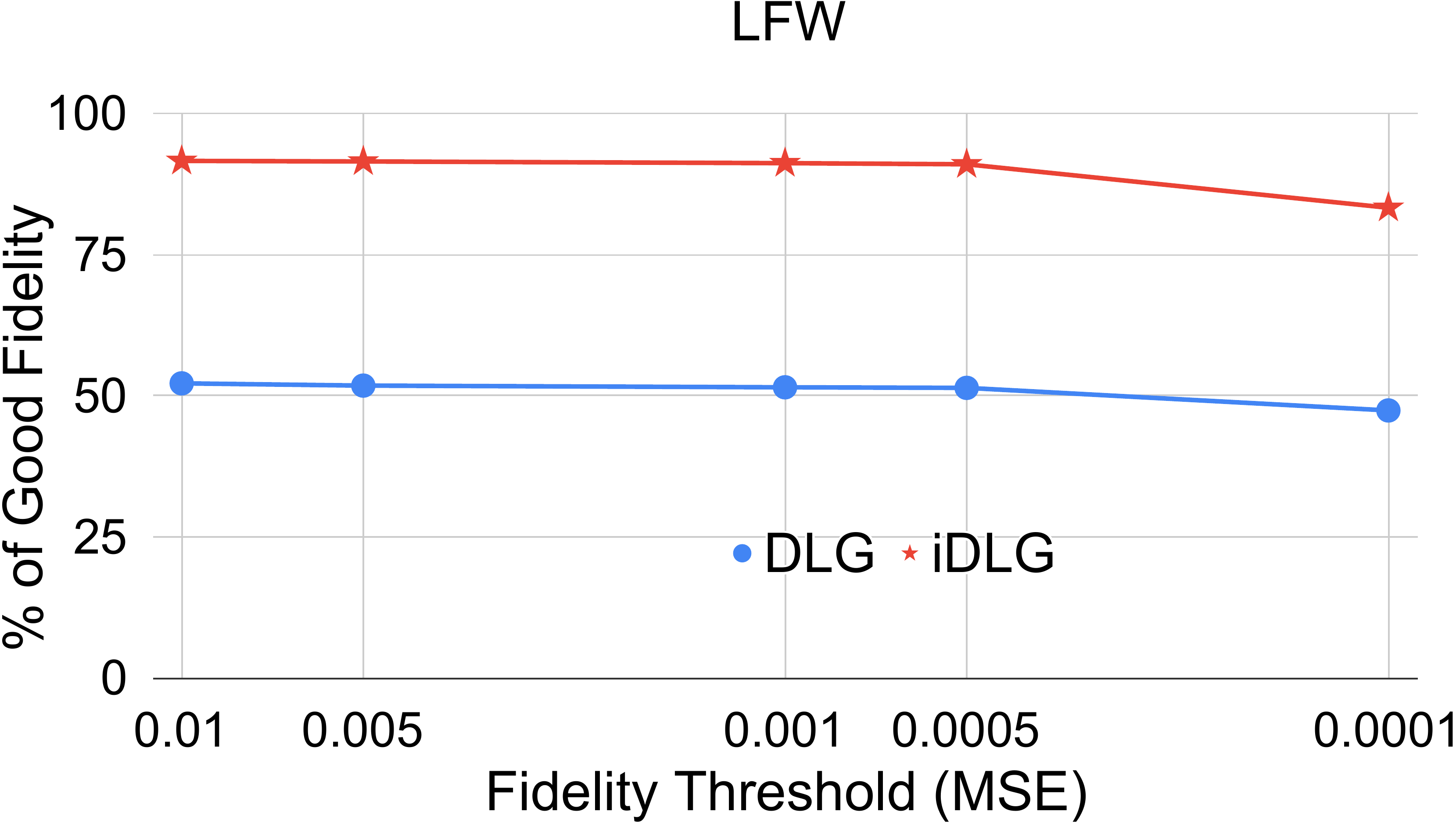}\hfill
  \vspace{0.01\textwidth}
\end{minipage}
\caption{\small{The Fidelity comparison of DLG~\cite{zhu19deep} and iDLG on three datasets. The x-axis denotes the (MSE) threshold of good fidelity. From left to right, the threshold decreases and the fidelity requirement improves. Obviously, the proposed iDLG consistently outperforms DLG in recovering data with significant margin on three tasks. The advantage of iDLG is remarkable on the hard task of LFW.}
}
\label{fig:convergence} 
\end{figure}

\subsection{The Accuracy of Extracted Labels}
\label{subsec:label-accuracy}
Table~\ref{tab:label-accuracy} shows the accuracy of the two methods to recover the ground-truth labels. It is clear that iDLG always extracts the correct label as opposed to DLG which extracts wrong labels many times. Specifically, the accuracy of DLG on MNIST, CIFAR-100 and LFW is 89.9\%, 83.3\% and 79.1\% respectively, which shows that DLG suffers more on harder tasks.

\subsection{The Fidelity of Extracted Data}
 \label{subsec:convergence}
In this subsection, we compare the fidelity of two data extraction methods (DLG and iDLG) by calculating the MSE (mean square error) between the dummy and original data. We vary the (MSE) threshold of good fidelity.
Figure~\ref{fig:convergence} shows the fidelity comparison of two methods under different thresholds over three datasets. 

The plots show the percentage of extracting (or generating) data with good fidelity. The x-axis indicates the (MSE) threshold for good fidelity. For example, $0.001$ means that we consider it good fidelity when the MSE between dummy and original data is less than $0.001$. From left to right, the threshold decreases and the fidelity requirement improves. Obviously, the proposed iDLG consistently outperforms DLG in recovering data with significant margin on three tasks. The advantage of iDLG is remarkable on the hard task of LFW. 

\begin{figure}[]
\centering
\noindent\begin{minipage}{\textwidth}
  \centering
  \includegraphics[width=.49\textwidth]{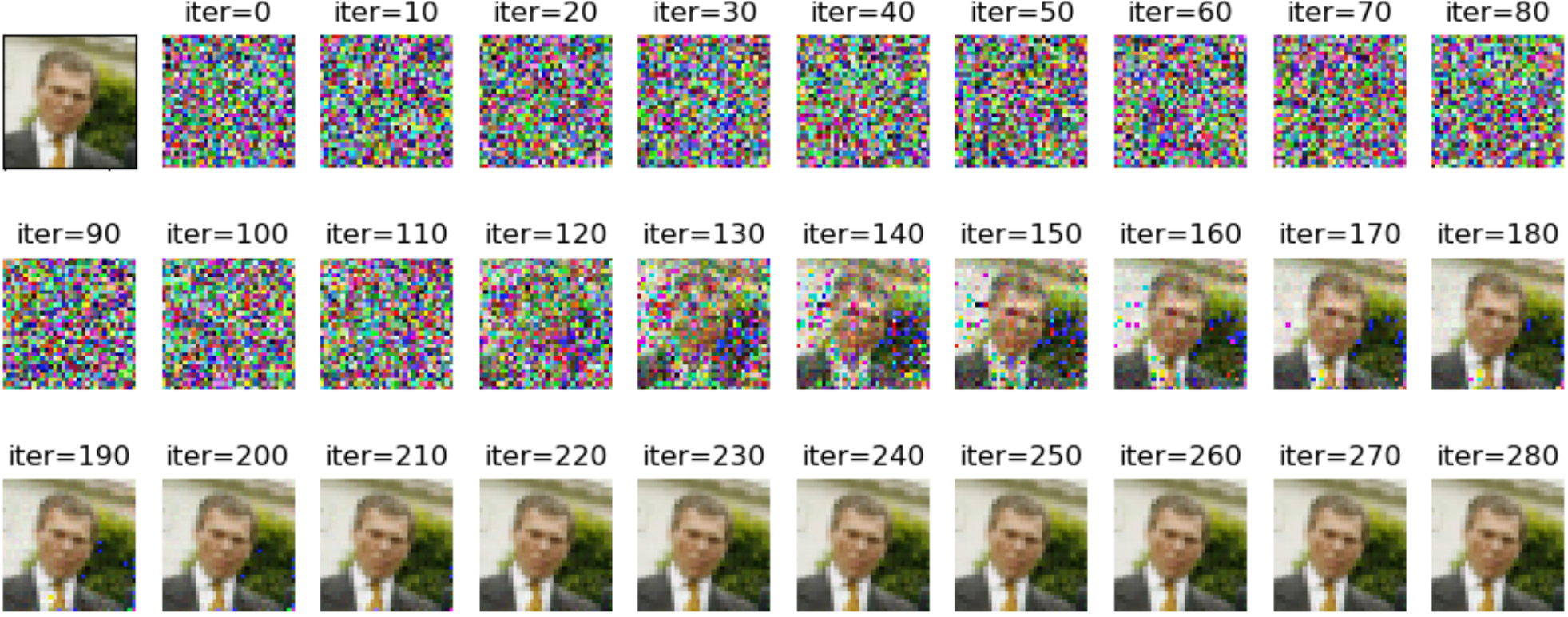}\hfill
  \includegraphics[width=.49\textwidth]{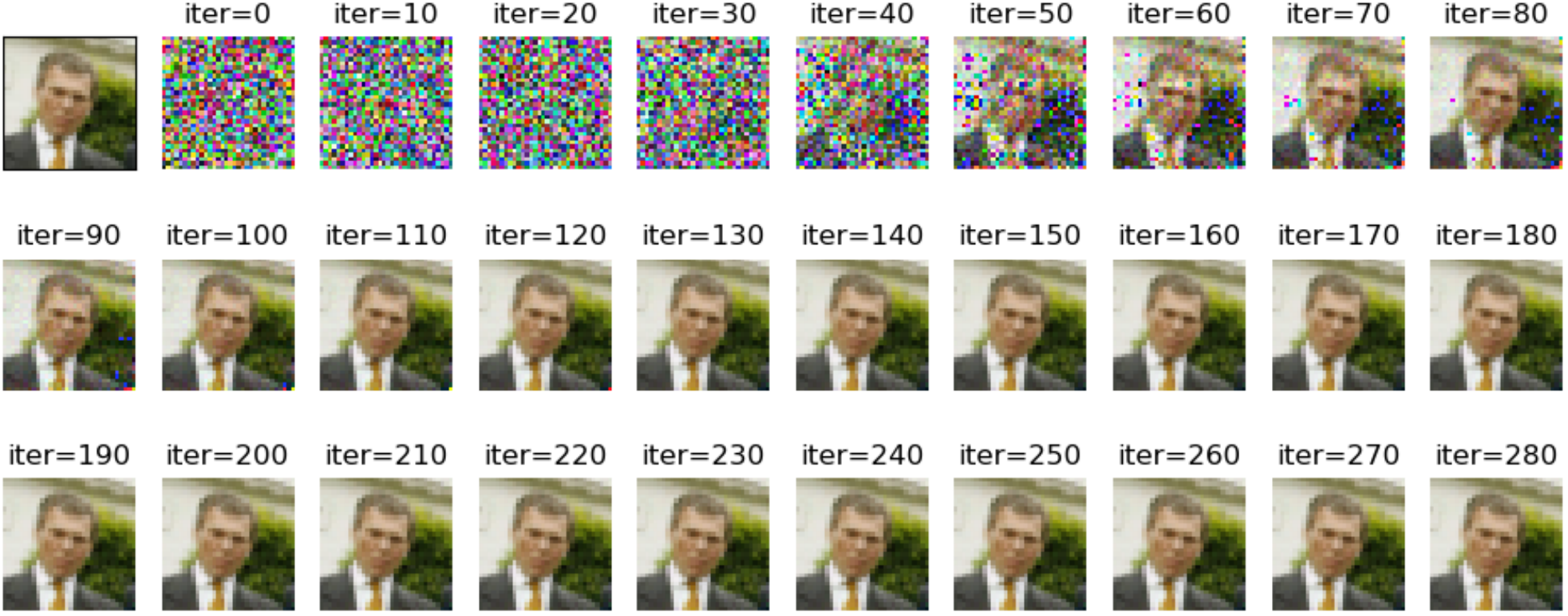}\hfill
\end{minipage}
\caption{\small{
Example of the training process of DLG (left) and iDLG (right) on LFW face dataset. The first image is the (original) private training image, and the followings are the extracted images at different training iterations. It is clear that the training of iDLG is easier to converge.}}
\label{fig:leaking} 
\end{figure}

Figure \ref{fig:leaking} gives an example of the training process of DLG (left) and iDLG (right) on LFW face dataset. The first image is the (original) private training image. The followings are the extracted images in different training iterations. It is clear that the training of iDLG is easier to converge. iDLG needs only 90 training iterations  to get the similar performance which requires DLG to train for 200 iterations.

\section{Discussion and Conclusion}
\label{sec:conclusion}
In this paper, we present an effective approach to steal the data and the corresponding labels from the shared gradients in a distributed training scenario. Particularly, we analytically illustrate the relationship between the labels and the signs of corresponding gradients. Based on this, our approach can extract the ground-truth labels with $100\%$ accuracy which facilitates the data extraction with increased fidelity. Currently, our method works with a simplified scenario of sharing gradients of every datum. In other words, iDLG can identify the ground-truth labels only if gradients w.r.t. every sample in a training batch are provided.

\small
\bibliographystyle{unsrt}
\bibliography{egbib}

\end{document}